\documentclass{article}

\usepackage{natbib}

\usepackage{amsmath,amsfonts,bm}

\def\eqref#1{equation~\ref{#1}}

\def\1{\bm{1}}

\DeclareMathAlphabet{\mathsfit}{\encodingdefault}{\sfdefault}{m}{sl}
\SetMathAlphabet{\mathsfit}{bold}{\encodingdefault}{\sfdefault}{bx}{n}

\newcommand{\softmax}{\mathrm{softmax}}

\DeclareMathOperator*{\argmax}{arg\,max}

 \usepackage[dblblindworkshop, final]{neurips_2026}

\usepackage[utf8]{inputenc} 
\usepackage[T1]{fontenc}    
\usepackage{hyperref}       
\usepackage{url}            
\usepackage{booktabs, graphicx, subcaption, multirow, enumitem}
\usepackage{amsfonts}       
\usepackage{nicefrac}       
\usepackage{microtype}      
\usepackage{xcolor}         
\usepackage{wrapfig, float, algorithm, algorithmic, amsmath, amssymb}

\title{Predictor-Guided Latent Space Codon Optimization for Maximizing Protein Expression}

\workshoptitle{AI for Drug Discovery: Bridging the Translation Gap}

\author{%
  \textbf{Alberto Caron}$^{1}$ \quad
  \textbf{Tianyu Cui}$^{1}$ \quad
  \textbf{Dmytro S.\,Lituiev}$^{1}$ \quad
  \textbf{Mangal Prakash}$^{2}$ \\
  \textbf{Artem Moskalev}$^{3}$ \quad
  \textbf{Amina Mollaysa}$^{1}$ \quad
  \textbf{Bo Zhai}$^{1}$ \quad
  \textbf{Hirsh Nanda}$^{1}$ \\
  \textbf{Daniel M.\,Poole}$^{1}$ \quad
  \textbf{Zhongyin Liu}$^{1}$ \quad
  \textbf{Iman Farasat}$^{1}$ \quad
  \textbf{Robert Davidson}$^{1}$ \\
  \textbf{Nikolay V.\,Manyakov}$^{1}$ \quad
  \textbf{Tommaso Mansi}$^{1}$ \quad
  \textbf{Scott Oloff}$^{1}$ \quad  
  \textbf{Rui Liao}$^{1}$ \\[0.5ex]
  \small $^{1}$Johnson \& Johnson Innovative Medicine \\
  \small $^{2}$Ellison Institute of Technology Oxford \quad $^{3}$New Theory AI
}

\begin{document}

\maketitle

\begin{abstract}
Codon optimization, the process of selecting synonymous codons to improve mRNA translation efficiency and protein expression, is central to therapeutic protein production and mRNA vaccines, yet it remains a hard problem. The design space is discrete and combinatorially large, precluding gradient-based methods, and existing tools rely on heuristic proxies (e.g., Codon Adaptation Index or GC-content) that poorly capture true expression. We introduce Latent-Space Codon Optimization (LSCO), which recasts this discrete problem as a continuous one by mapping sequences into the latent space of a pretrained mRNA language model, enabling efficient gradient-based search. LSCO combines four components: a data-driven expression objective from an uncertainty-aware predictor, a Minimum-Free-Energy regularizer for structural stability, a naturalness prior from a protein-to-codon back-translation model, and constrained decoding for protein fidelity. On a real-world, wet-lab antibody expression dataset, LSCO outperforms simple frequency-based, as well as modern deep generative baselines in predicted expression, while retaining suitable biophysical properties.
\end{abstract}

\section{Introduction}

Synonymous codons, i.e., nucleotide triplets that redundantly encode the same amino acid, create a vast design space for tailoring mRNA without changing the associated protein~\citep{brule2017synonymous}. This space grows combinatorially with protein length: a therapeutic antibody such as trastuzumab (made of a 449 and 214 amino acid heavy and light chains, respectively) admits on the order of $10^{300}$ synonymous encodings~\citep{Carter1992}. Codon choice within this space strongly affects translation efficiency and protein yield, so codon optimization is a key lever in applications from monoclonal antibody (mAb) manufacturing to mRNA therapeutics~\citep{mauro2014critical, paremskaia2024codon}.


Codon optimization faces two challenges. First, the synonymous space is discrete and combinatorially vast, ruling out exhaustive search and gradient-based methods. Second, the true objective, an accurate mRNA-to-protein expression model, is inaccessible: the biology is only partly understood and labelled data is scarce, forcing reliance on heuristic proxies (CAI, GC-content, codon-pair bias, ``naturalness'') that enable tractable algorithms, but often fail to capture true protein expression~\citep{sharp1987codon, constant2023deep}. Note that, while the optimization objective is formulated in terms of protein expression, experimental validation typically uses A280 absorbance as a practical proxy for protein yield and abundance \citep{pace1995measure}. To address both challenges, we propose Latent-Space Codon Optimization (LSCO), which recasts codon optimization as a continuous, gradient-based problem grounded in a data-driven expression objective. LSCO optimizes a continuous latent representation derived from a pretrained mRNA language model, bypassing discrete search, and integrates four components: i) an expression objective from an uncertainty-aware predictor trained on mRNA–protein data that penalizes out-of-distribution drift during optimization; ii) a Minimum Free Energy (MFE) regularizer that avoids near-adversarial, unstable sequences; iii) a naturalness prior from a protein-to-codon back-translation model, learned from the Observed Antibody Space (OAS) data~\citep{olsen2022observed}, that keeps candidates within plausible codon usages; iv) constrained decoding that ensures protein fidelity. Our contributions can be summarized as follows:
\begin{itemize}[leftmargin=0.8em, topsep=0pt, itemsep=0pt]
    \item We propose LSCO, a continuous latent-space codon optimizer featuring the four novel components listed above that replaces heuristic proxies with reliable expression surrogate as main objective.
    \item On a real-world, wet-lab antibody expression dataset, LSCO outperforms frequency-based and modern deep generative baselines on predicted expression, as confirmed by independent held-out predictors, while retaining structural stability and host-appropriate GC content.
    \item Ablations show the MFE regularizer offers a controllable expression–stability trade-off, and both the pretrained latent space (vs.\ a direct input-space search) and the naturalness prior drive the gains.
\end{itemize}

\section{Background and Related Work}
\label{sec:background}

\paragraph{Problem Setup.}
A protein sequence of length $L$ is denoted by $\mathbf{a} = (a_1, \ldots, a_L)$, where $a_i \in \mathcal{A}$ are the 20 canonical amino acids. Each amino acid $a_i$ is encoded by many synonymous mRNA sequences $\mathbf{c} = (c_1, \ldots, c_L)$, with $c_i \in \mathcal{C}(a_i)$ being the synonymous codons for $a_i$. We represent codons in one-hot form $R \in [0,1]^{L\times 64}$, where $\mathbf{r}_i$ is the one-hot vector encoding of codon $c_i$. Codons can be viewed as drawn from a $c_i \sim \text{Categorical}(\tilde{\mathbf{r}}_i)$ distribution, with probabilities $\tilde{\mathbf{r}}_i$. Codon optimization searches the combinatorial design space of codons mapping to the same protein, $\mathcal{S}(\mathbf{a}) = \mathcal{C}(a_1)\times\cdots\times\mathcal{C}(a_L)$, for an mRNA sequence $\mathbf{c}^\star$ that maximizes downstream protein expression,
    \begin{equation}
    \label{eq:codopt_formulation}
        \mathbf{c}^\star = \argmax_{\mathbf{c} \in \mathcal{S}(\mathbf{a})} \Phi (\mathbf{c}),
    \end{equation}
where $\Phi$ is a surrogate mRNA-to-protein expression model. 
Both the inaccessibility of $\Phi$ (underlying biology is only partly understood and labelled data is scarce) and the discrete, combinatorial nature of $\mathcal{S}(\mathbf{a})$ make this problem intractable for exhaustive or gradient-based search~\citep{moore2002codonopt, arbib2020codon}.

\paragraph{Related Work.}
Codon optimization has evolved from frequency-based tools (JCat~\citep{grote2005jcat}, OPTIMIZER~\citep{puigbo2007optimizer}) that maximize CAI and codon-pair bias but neglect determinants such as mRNA stability~\citep{zhang2023lineardesign, paremskaia2024codon}, to deep generative mRNA language models pre-trained to learn codon usage from large scale datasets without optimizing any sequence properties post-hoc (CodonBERT~\citep{li2024codonbert}, CodonTransformer~\citep{fallahpour2025codontransformer}, GEMORNA~\citep{zhang2025deep}) and per se multi-objective post-hoc sequence optimizers (RNop~\citep{gong2025new}, ICOR~\citep{jain2023icor}).
Despite advances in sequence modeling and optimization, these methods generally optimize surrogate sequence features rather than protein expression, creating a potential gap between optimization objectives and the desired biological outcome.

General-purpose methods for optimizing discrete sequences (not just codon optimization) differ in \emph{search space geometry} (\emph{where} they search) and \emph{trust mechanism} (\emph{how} they avoid ``reward hacking", i.e., exploiting an unreliable surrogate off-distribution). \emph{Sequence-space} methods operate directly on sequences or distributions over them~\citep{deb2002fast, brookes2019dbas, trabucco2021conservative, angermueller2020dynappo}, and evolve candidates by mutation, selection and screening \citep{deb2002fast, yang2019machine, zhang2021unifying}; whereas \emph{latent-space} methods operate in a learned continuous space shared by a decoder and a property prediction model~\citep{gomez2018automatic, tripp2020sample}, with latent-space Bayesian optimization adding a probabilistic surrogate for uncertainty-aware search~\citep{stanton2022accelerating, gruver2023protein, amin2024bayesian}. We include an extended version of related work in Appendix~\ref{app:extended_related}.

LSCO belongs to the \emph{latent-space} family and introduces three novel components: 
(i) it enforces realistic codon-usage priors using a mRNA foundation autoencoder model pretrained on a large scale dataset;
(ii) it constructs a meaningful expression surrogate by leveraging a predictor trained on a domain-specific dataset, instead of relying on heuristic proxies, and embeds this within a multi-criteria optimization objective that further balances stability, naturalness and protein fidelity;
(iii) it penalizes out-of-distribution drift during optimization via epistemic uncertainty estimates obtained from a deep ensemble version of the expression surrogate predictive model.


\begin{figure}[!ht]
    \centering
    \includegraphics[width=0.55\linewidth]{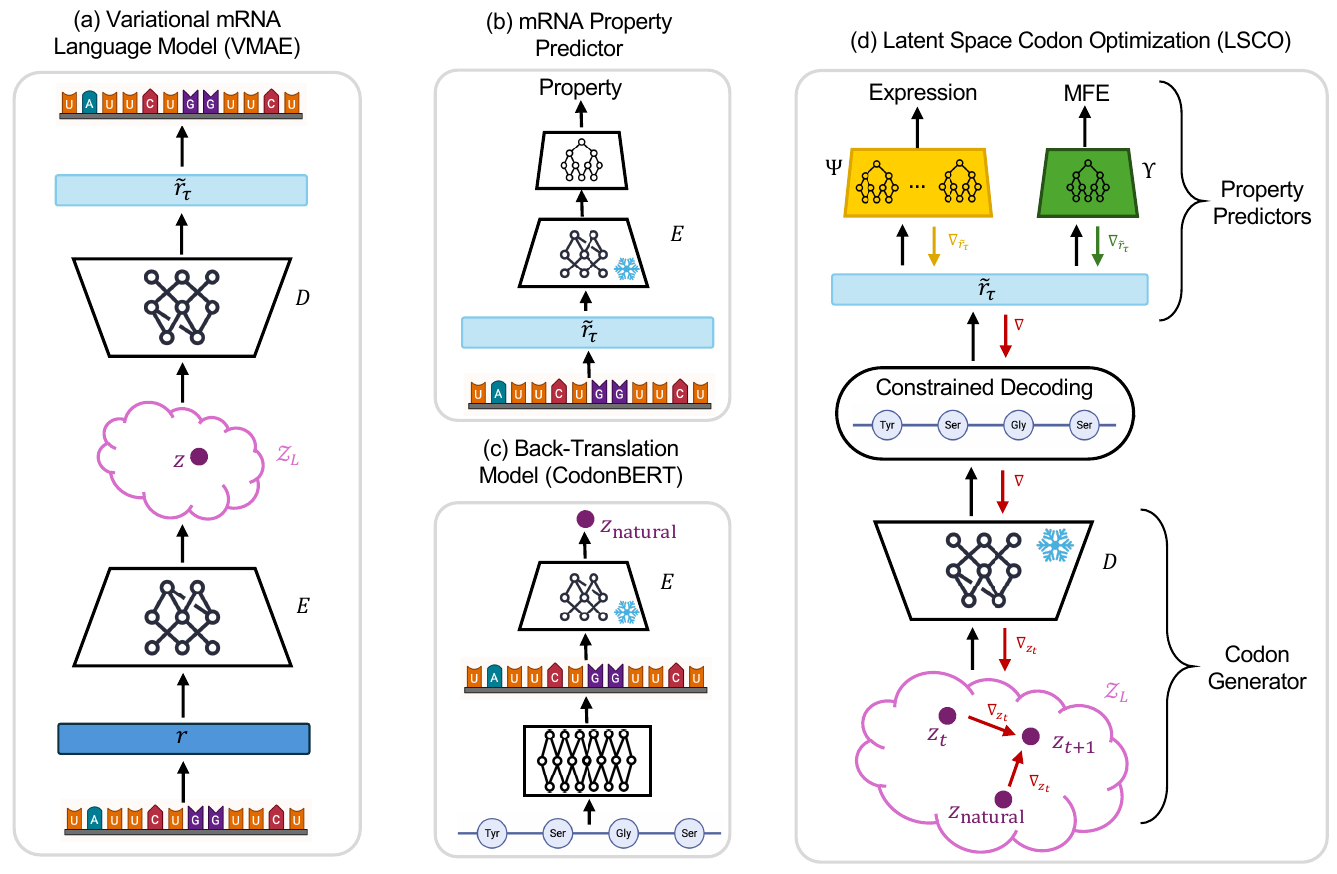}
    \caption{
        \textbf{LSCO framework.} 
        \textbf{(a)} mRNA sequences are mapped to codon-latent space using a pretrained variational masked language model. 
        \textbf{(b)} mRNA property predictor reuses the pretrained encoder in (a) and takes the soft and continuous version of the discrete sequence. 
        \textbf{(c)} We generate a natural mRNA sequence using a pretrained protein-to-codon back-translation model, whose representation $\mathbf{z}_{\text{nat}}$ from the pretrained encoder in (a) is used as the naturalness prior.
        \textbf{(d)} Latent representations $\mathbf{z}$ are optimized with respect to a data-driven expression predictor $\Psi$, an MFE predictor $\text{MFE}$, and a naturalness prior $\mathbf{z}_{\text{nat}}$ from (c). Constrained decoding ensures valid codon choices, preserving the target protein sequence. The result is an optimized mRNA that balances expression and stability.
    }
    \label{fig:method}
\end{figure}

\section{Latent-Space Codon Optimization}
\label{sec:lsco}


We propose Latent-Space Codon Optimization (LSCO) to address the challenges of codon optimization. First, LSCO recasts the discrete search into a continuous one: rather than searching the mRNA sequences space directly, we optimize codon-level embeddings in the latent space of a pretrained mRNA language model, then decode them back to the discrete sequence space. Second, instead of relying on heuristic proxies such as CAI or naturalness, we train a model $\Psi: [0,1]^{L\times 64} \rightarrow \mathbb{R}$ to predict mRNA-to-protein yield, and use a utility over its predictions, $U_{\Psi}$--an uncertainty-aware, saturating transform of $\Psi$--as main objective. Since optimizing expression alone can yield near-adversarial sequences with poor stability or GC content, we add two regularizers: a predicted Minimum Free Energy term, ${\text{MFE}}: [0,1]^{L\times 64} \rightarrow \mathbb{R}$, and a naturalness prior $U_{\text{nat}}$, together producing sequences that are highly expressed, stable, and within a host-appropriate GC range. The overall LSCO framework is depicted in Figure~\ref{fig:method}. Given a target protein $\mathbf{a}$, the LSCO objective is
\begin{gather} 
\label{eq:obj_lat}
    \hat{\mathbf{z}} = \argmax_{\mathbf{z}} \Big \langle \lambda_{\Psi} \cdot U_{\!\Psi}\!\!\left(D(\mathbf{z})\right) 
    - \lambda_\text{MFE}\cdot \text{MFE}\!\left(D(\mathbf{z})\right) + \lambda_{\text{nat}} \cdot U_{\text{nat}}(\mathbf{z}) \Big \rangle \quad
    \textit{s.t.} \quad D(\mathbf{z}) \in \mathcal{S}(\mathbf{a}),
\end{gather}
where positive $\lambda_{\Psi}, \lambda_\text{MFE}, \lambda_{\text{nat}}$ re-weight expression, MFE, and the naturalness prior; $D$ is the latent-to-sequence decoder (generator); and $\mathcal{S}(\mathbf{a})$ is the synonymous codons space for given protein $\mathbf{a}$. 
LSCO builds on four main components (Figure~\ref{fig:method}; full details in Appendix~\ref{appsec: implementation})

\textbf{i) Variational mRNA language model:} a bidirectional transformer we pretrain on the curated OAS-mRNA dataset~\citep{yazdani2025helm} (15M+ sequences) with a Variational Masked Auto-Encoder objective~\citep{hu2022varmae}. Its encoder $E: \mathbb{R}^{L\times64}\!\!\rightarrow\!\mathcal{Z}_L$ maps codons to a continuous latent space $\mathcal{Z}_L$, while its decoder $D: \mathcal{Z}_L \!\rightarrow\!\mathbb{R}^{L\times64}$ reconstructs codons through a temperature-$\tau$ softmax, that can yield a continuous relaxation $\tilde{\mathbf{r}}_{\tau}$ of the codon sequence that is a smooth, differentiable surrogate for the discrete mRNA. 

\textbf{ii) Expression predictor:} a deep learning model $\Psi: [0,1]^{L\times64} \rightarrow \mathbb{R}$, built on top of the pretrained encoder $E$ and trained on mRNA–protein expression data, that takes $\tilde{\mathbf{r}}_{\tau}$ as inputs rather than $\mathcal{Z}_L$ (which keeps prediction tied to the sequence and enables constrained decoding). We implement $\Psi$ as a deep ensemble~\citep{lakshminarayanan2017simple} of $M{=}10$ members (Appendix~\ref{app:deepens}), which outputs an ensemble mean $\mu_\Psi$ and standard deviation $\sigma_\Psi$. The expression-related optimization utility in Eq.~\ref{eq:obj_lat} is a Lower Confidence Bound (LCB) constructed on these predictions
\begin{equation}
\label{eq:expression_utility}
U_{\Psi}(\tilde{\mathbf{r}}_{\tau}) = \mu_{\Psi}(\tilde{\mathbf{r}}_{\tau}) - \lambda_{\text{LCB}}\, \sigma_{\Psi}(\tilde{\mathbf{r}}_{\tau}),
\end{equation}
which maximizes protein yield while penalizing high-uncertainty, out-of-distribution designs.

\textbf{iii) Minimum Free Energy (MFE) regularizer:} since optimizing expression alone in Eq.~\ref{eq:obj_lat} might yield unstable sequences, we add a regularizing terms based on a predictor $\text{MFE}: [0,1]^{L\times64} \rightarrow \mathbb{R}$ (similar architecture as $\Psi$, no ensemble), trained on ViennaRNA-computed MFE  labels~\citep{lorenz2011viennarna}; lower MFE implies more stable structure~\citep{higgs2000rna}. 

\textbf{iv) Naturalness prior:} We pretrain a CodonBERT back-translation model~\citep{devlin2019bert, li2024codonbert} on the OAS-mRNA data. This model learns a natural conditional codon distribution $p(\mathbf{c}|\mathbf{a})=\prod_i p(c_i|\mathbf{a})$ and produces a plausible reference mRNA sequence $\mathbf{c}_{\text{nat}} = \argmax_{\mathbf{c}} p(\mathbf{c}|\mathbf{a})$. We add a naturalness-based regularizer that penalizes sequences far from $\mathbf{z}_{\text{nat}}$ in the latent-space to keep them within biologically plausible codon usage $U_{\text{nat}}(\mathbf{z}) = -\|E(\mathbf{c}) - E(\mathbf{c}_{\text{nat}})\|_2 = -\|\mathbf{z} - \mathbf{z}_{\text{nat}}\|_2$.

Finally, \textbf{constrained decoding} is implemented to guarantee protein fidelity: a per-position validity mask sets non-synonymous codon logits to $-\infty$ before applying temperature-$\tau$ softmax, in order to enforce $\tilde{\mathbf{r}}_{\tau} \in \mathcal{S}(\mathbf{a})$ while the optimizer explores synonymous codon choices. More details about LSCO implementation and a LSCO pseudocode (Algorithm~\ref{alg:lsco}) are provided in Appendix~\ref{appsec: implementation}.

\section{Experiments}
\label{sec:experiments}

\begin{figure}[t!]
    \centering
    \includegraphics[width=0.83\linewidth]{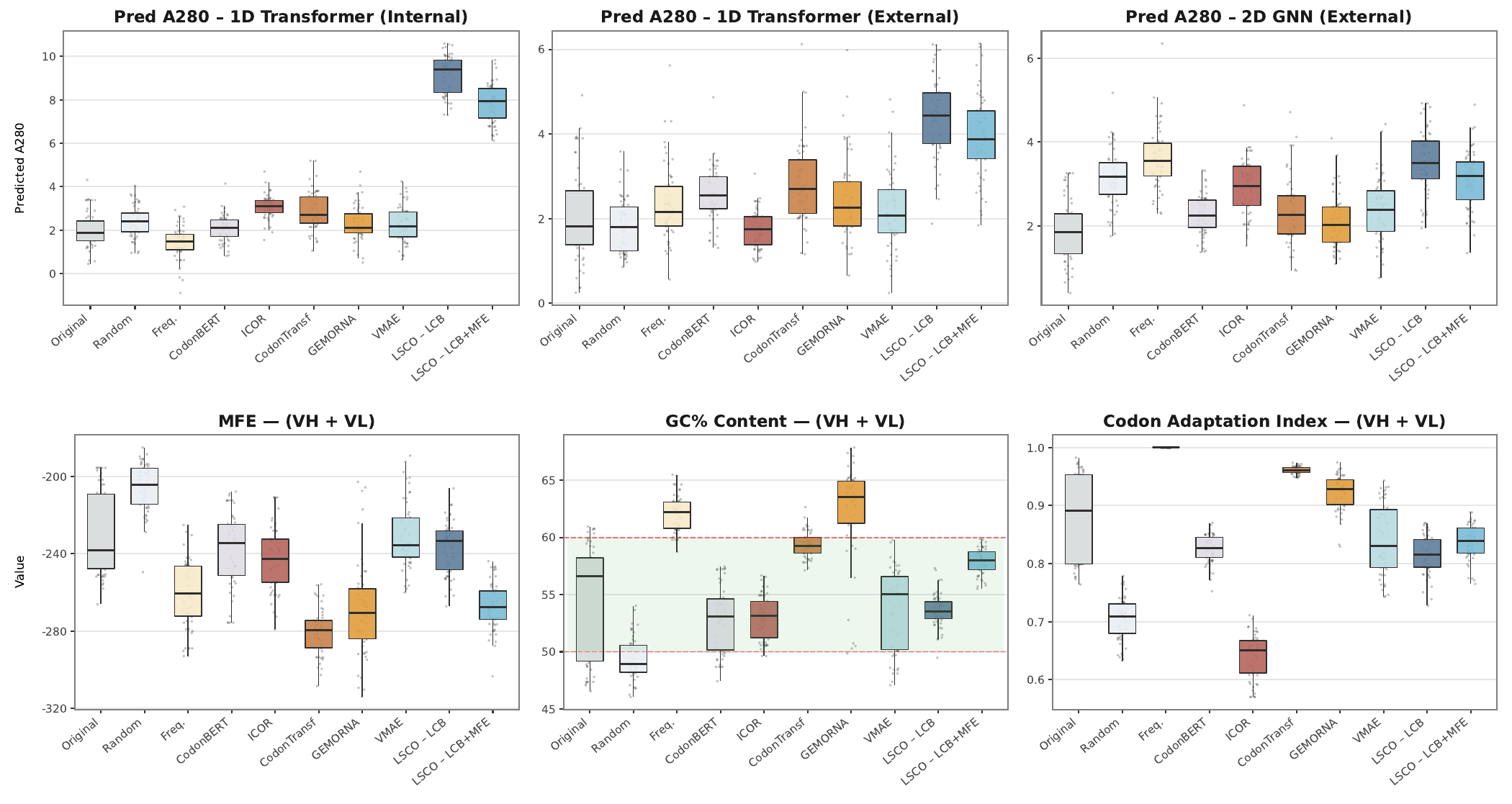}
    \caption{\textbf{Optimized sequence comparison on 54 mAbs.}
\emph{Top row:} predicted A280 from three models: LSCO's native 1D transformer, an external 1D transformer, and an external structure-aware 2D GNN. LSCO variants attain the highest or comparable expression across all three.
\emph{Bottom row:} biophysical metrics on heavy + light chains: MFE (lower = more stable), GC\% (green band = host-appropriate range), CAI. \textit{LSCO-LCB+MFE} variant recovers stability and nudges GC/CAI upward at a modest expression cost, staying within host-appropriate ranges.}    
\label{fig:main_results}
\end{figure}

We benchmark LSCO against other methods from three different families: i) frequency-based, host-preferred selection~\citep{sharp1987codon}; ii) a maximum-naturalness CodonBERT back-translation model~\citep{li2024codonbert}; and iii) modern deep generative models/optimizers, including ICOR~\citep{jain2023icor}, CodonTransformer~\citep{fallahpour2025codontransformer}, and GEMORNA~\citep{zhang2025deep}. Random synonymous codon substitution and the Variational Masked Auto-Encoder (VMAE) generator (without optimizing for predicted expression) are included to serve as lower-bound references. 
We assess generated mRNA sequences using four criteria: \textit{protein yield}, measured via predicted absorbance at 280 nm (A280) from our native 1D predictor $\Psi$ and two independent held-out 1D and 2D predictors; \textit{thermodynamic stability} via Minimum Free Energy (MFE) computed via the ViennaRNA package~\citep{lorenz2011viennarna}); \textit{host compatibility} via GC content (GC\%)~\citep{kudla2009coding}; and \textit{codon adaptation} via Codon Adaptation Index (CAI), reported as a conventional, but coarse, proxy for expression.

\paragraph{LSCO Evaluation.}

We evaluate LSCO and baselines on 54 starting monoclonal antibodies (mAbs), selected based on their potential margin of improvement and stratified by target antigen (full details on the selection in Appendix~\ref{appsec: implementation}). We score every optimized design by predicted A280 with the native 1D predictor and the two \emph{independent} held-out evaluators, an external 1D transformer with a different backbone and a structure-aware 2D GNN. We test two LSCO variants: \textit{LSCO--LCB}, optimizing only expression $U_\Psi$ and naturalness $U_{\text{nat}}$ (with $\lambda_\text{MFE}=0$), and \textit{LSCO--LCB+MFE}, which adds MFE regularization ($\lambda_\text{MFE}=0.1$ unless otherwise specified). As shown in the top row of Figure~\ref{fig:main_results}, LSCO attains the highest predicted A280 across both 1D internal and external predictors and scores comparably on the 2D GNN evaluator. The general agreement among the predictors, despite their different inductive biases, indicates the gains are not just an artifact of the optimizer exploiting its own surrogate, but reflect genuinely higher predicted expression as a direct benefit of optimizing expression in a continuous latent space rather than proxy surrogates. 
The bottom row shows the stability/host-compatibility trade-off. The expression-only \textit{LSCO--LCB} variant maximizes A280 but delivers weaker MFE than strong baselines like CodonTransformer and GEMORNA. Adding the MFE regularizer (\textit{LSCO--LCB+MFE}) recovers stability by driving MFE substantially lower and nudging GC and CAI upward, yet well within host-appropriate ranges, at a modest expression cost. 

\begin{table*}[t]
\centering

\resizebox{0.92\textwidth}{!}{%
\begin{tabular}{clccc ccc ccc}
\toprule
& & \multicolumn{3}{c}{Configuration}
& \multicolumn{3}{c}{Predicted A280}
& \multicolumn{3}{c}{Metrics (VH + VL)} \\
\cmidrule(lr){3-5}
\cmidrule(lr){6-8}
\cmidrule(lr){9-11}

Study & Space
& $\lambda_{\mathrm{MFE}}$
& $\lambda_{\mathrm{LCB}}$
& $\lambda_{\mathrm{nat}}$
& 1D Internal
& 1D External
& 2D External
& MFE
& GC\%
& CAI \\
\midrule

\multirow{6}{*}{\shortstack{MFE\\Sensitivity}}
& \multirow{6}{*}{LSCO}
& 0   & 0 & 0.01
& $\mathbf{9.585 \pm 0.218}$
& $\mathbf{4.518 \pm 0.223}$
& $\mathbf{3.492 \pm 0.203}$
& $-235.187 \pm 3.555$
& $53.303 \pm 0.423$
& $0.816 \pm 0.008$ \\

&
& 0   & 1 & 0.01
& $\mathbf{9.156 \pm 0.241}$
& $\mathbf{4.400 \pm 0.254}$
& $\mathbf{3.485 \pm 0.212}$
& $-236.711 \pm 3.603$
& $53.601 \pm 0.390$
& $0.815 \pm 0.009$ \\

&
& 0.1 & 0 & 0.01
& $8.180 \pm 0.237$
& $3.900 \pm 0.224$
& $3.175 \pm 0.217$
& $-266.274 \pm 2.783$
& $57.756 \pm 0.285$
& $0.830 \pm 0.009$ \\

&
& 0.1 & 1 & 0.01
& $7.858 \pm 0.235$
& $3.935 \pm 0.265$
& $3.099 \pm 0.187$
& $\mathbf{-267.050 \pm 3.211}$
& $57.916 \pm 0.300$
& $0.837 \pm 0.008$ \\

&
& 1   & 0 & 0.01
& $8.065 \pm 0.232$
& $3.887 \pm 0.200$
& $3.319 \pm 0.201$
& $-263.765 \pm 3.122$
& $58.003 \pm 0.304$
& $0.825 \pm 0.009$ \\

&
& 1   & 1 & 0.01
& $7.899 \pm 0.256$
& $3.951 \pm 0.271$
& $2.923 \pm 0.182$
& $\mathbf{-266.365 \pm 3.381}$
& $58.018 \pm 0.317$
& $\mathbf{0.838 \pm 0.008}$ \\

\midrule

\multirow{2}{*}{\shortstack{Naturalness\\Ablation}}
& \multirow{2}{*}{LSCO}
& 0   & 1 & 0
& $9.024 \pm 0.247$
& $\mathbf{4.564 \pm 0.240}$
& $3.248 \pm 0.193$
& $-236.491 \pm 3.900$
& $53.428 \pm 0.465$
& $0.808 \pm 0.008$ \\

&
& 0.1 & 1 & 0
& $7.956 \pm 0.208$
& $3.885 \pm 0.249$
& $3.259 \pm 0.254$
& $-265.006 \pm 3.066$
& $57.860 \pm 0.305$
& $0.836 \pm 0.009$ \\

\midrule

\multirow{4}{*}{\shortstack{Search-Space\\Ablation}}
& \multirow{4}{*}{DISO}
& 0   & 1 & 0.01
& $8.547 \pm 0.314$
& $4.423 \pm 0.240$
& $3.220 \pm 0.192$
& $-239.578 \pm 3.904$
& $53.594 \pm 0.461$
& $0.829 \pm 0.007$ \\

&
& 0   & 1 & 0
& $8.528 \pm 0.305$
& $4.393 \pm 0.230$
& $3.036 \pm 0.174$
& $-239.561 \pm 3.678$
& $53.688 \pm 0.421$
& $0.829 \pm 0.007$ \\

&
& 0.1 & 1 & 0.01
& $7.631 \pm 0.268$
& $3.828 \pm 0.241$
& $2.939 \pm 0.206$
& $-264.987 \pm 3.625$
& $57.428 \pm 0.353$
& $0.839 \pm 0.010$ \\

&
& 0.1 & 1 & 0
& $7.598 \pm 0.246$
& $3.840 \pm 0.247$
& $2.923 \pm 0.187$
& $-264.994 \pm 3.639$
& $57.382 \pm 0.355$
& $\mathbf{0.840 \pm 0.010}$ \\

\bottomrule
\end{tabular}%
}
\caption{\textbf{Sensitivity and ablation studies.}
Predicted protein yield (A280) from the internal 1D Transformer and two held-out
evaluators, together with biophysical metrics for the joint heavy and
light chains, on the same 54 starting mAbs as
Figure~\ref{fig:main_results}. Values are means with 95\% confidence
bands. LSCO searches in the pretrained VMAE latent space, whereas DISO
optimizes codon logits directly in input space. The
$\lambda_{\mathrm{MFE}}$ and $\lambda_{\mathrm{LCB}}$ columns respectively
control MFE regularization and the uncertainty penalty; $\lambda_{\mathrm{nat}}=0$ denotes
removal of the naturalness prior.
Overall, $\lambda_{\mathrm{MFE}}=0.1$ provides good
expression--stability trade-off, and LSCO generally outperforms DISO.}
\label{tab:combined_sensitivity_ablation}
\end{table*}

\textbf{Ablation and Sensitivity Studies.}
We run two additional experiments on LSCO components: a sensitivity analysis sweeping over different MFE weights, and an ablation study aimed at isolating the effect of pretrained VMAE latent space search vs.\ direct input-space search. Both report predicted A280 from the three predictors plus MFE, GC\%, and CAI.

The \emph{MFE Sensitivity} block in Table~\ref{tab:combined_sensitivity_ablation} sweeps $\lambda_{\text{MFE}}$ regularization weight while toggling $\lambda_{\text{LCB}}$, which sets the uncertainty penalty strength in Eq.~\ref{eq:expression_utility}. The expression-only variant ($\lambda_{\text{MFE}}=0$) gives the highest A280 across all predictors but the weakest stability and lowest GC. A light penalty ($\lambda_{\text{MFE}}=0.1$) drives MFE substantially lower and lifts GC and CAI into the host-appropriate range at a modest expression cost. Larger weights ($\lambda_{\text{MFE}}=1$) yield no further stability. Enabling the uncertainty penalty ($\lambda_{\text{LCB}}=1$) only slightly lowers A280 while keeping the optimization from drifting off-distribution. Thus, we set $\lambda_{\text{MFE}}=0.1, \lambda_{\text{LCB}} = 1$ for \textit{LSCO--LCB+MFE} (Figure~\ref{fig:main_results}).

The \emph{Search-Space Ablation} block of Table~\ref{tab:combined_sensitivity_ablation} compares LSCO against Direct Input-Space Optimization (DISO), which optimizes per-position codon logits directly, with no VMAE decoder in the loop, sharing all else with LSCO (details in Appendix~\ref{app:diso}). LSCO generally exceeds DISO on predicted A280 across evaluators and on biophysical metrics, showing that optimizing in the pretrained latent space is beneficial. Ablating the naturalness prior ($\lambda_{\text{nat}}=0$), corresponding to the \emph{Naturalness Ablation} block of Table~\ref{tab:combined_sensitivity_ablation}, leaves the native predictor's A280 largely unchanged but degrades the held-out 2D-GNN score—its expected signature as an on-manifold regularizer that keeps designs plausible to independent evaluators rather than inflating the optimizer's own objective.

\section{Conclusion}
\label{sec:discussion}

We introduced Latent-Space Codon Optimization (LSCO), which recasts codon optimization as a continuous, data-driven search in the latent space of a pretrained mRNA language model, steered by an uncertainty-aware expression predictor, MFE regularization, a naturalness prior, and constrained decoding to guarantee protein fidelity. On antibody data, LSCO outperforms frequency-based and modern deep generative methods on predicted expression from internal and two external models. Ablation and sensitivity studies demonstrate the benefits of balancing translation efficiency (A280) and stability (MFE) and operating in the pretrained VMAE latent space. As reliable mRNA-protein expression data are scarce, our evaluation relies on predicted rather than empirical yield. Future work will include wet-lab validation and design of a lab-in-the-loop Bayesian optimization framework.

\bibliographystyle{abbrv}
\bibliography{references}


\newpage
\appendix

\section{Extended Related Work}
\label{app:extended_related}

\begin{figure}[t]
    \centering
    \includegraphics[width=\linewidth]{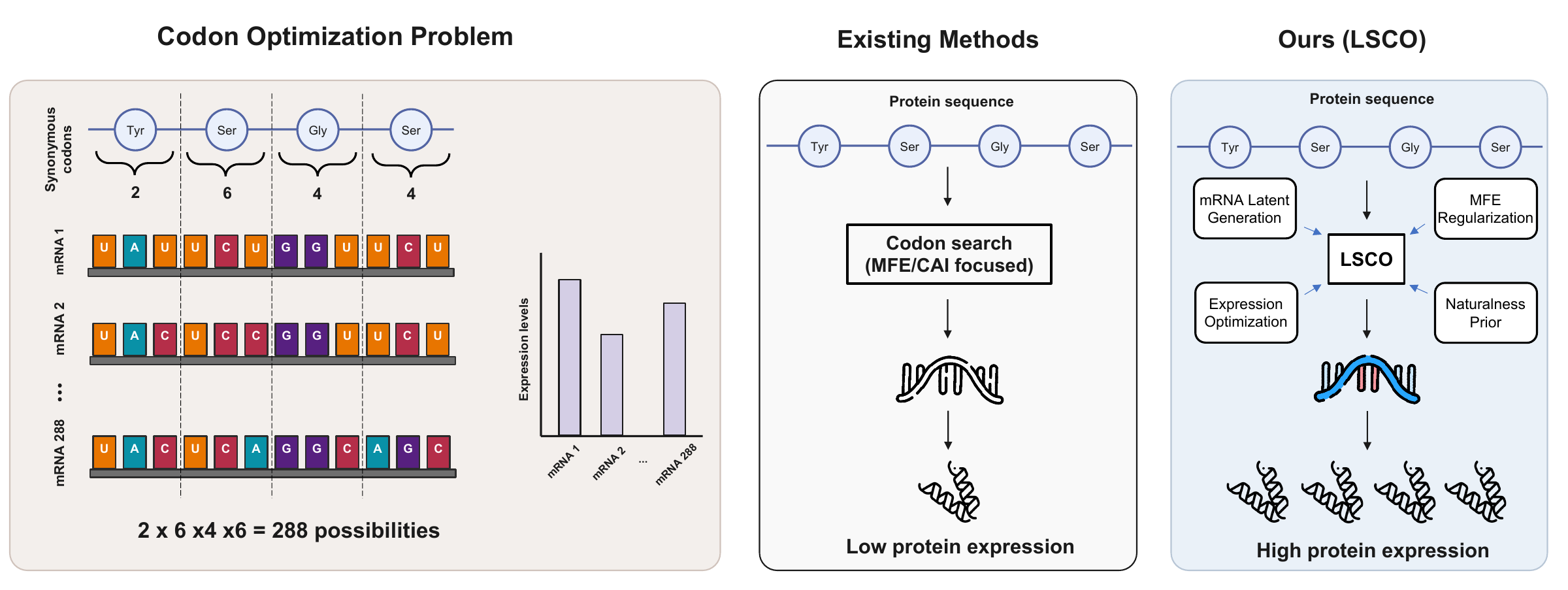}
    \caption{
        \textbf{Generative Codon optimization problem.} 
        Starting from a protein sequence, we navigate the combinatorial codon design space by mapping to and optimizing in a continuous latent space. LSCO integrates differentiable predictors for expression and stability, a naturalness prior, and constrained decoding to produce host-compatible, high-expression mRNA sequences.
    }
    \label{fig:teaser}
\end{figure}

\paragraph{Codon Optimization}
Codon optimization is a longstanding challenge in synthetic biology and biotechnology, aiming to boost recombinant protein expression by tailoring gene sequences to a host's codon preferences. A visual depiction of the codon optimization problem is provided in Figure \ref{fig:teaser}. Early strategies focused heavily on maximizing the use of highly frequent codons within the host organism, leading to tools like JCat~\citep{grote2005jcat} and OPTIMIZER~\citep{puigbo2007optimizer} that emphasize codon adaptation index (CAI) and codon pair bias as key metrics. However, such traditional approaches have shown limitations, including insufficient attention to other genomic features such as mRNA stability and resulting protein expression~\citep{zhang2023lineardesign, demissie2025comparative, paremskaia2024codon}.

Recent advances have seen the incorporation of deep learning into the codon optimization pipeline. Pretrained mRNA language models based on transformer architectures, such as CodonBERT~\citep{li2024codonbert}, CodonTransformer~\citep{fallahpour2025codontransformer} and GEMORNA~\citep{zhang2025deep}, leverage large-scale, multi-species datasets to learn complex codon usage patterns across diverse organisms via amino-acid–codon pair representations and novel tokenization, ensuring species specificity and preservation of regulatory elements. Similarly, RNop~\citep{gong2025new} employs a suite of custom loss functions that target fidelity, tRNA availability, mRNA secondary structure, and codon adaptation within a transformer architecture. Other recent work further leverages recurrent networks and active learning to improve translational efficiency across expression systems~\citep{jain2023icor, li2024deep}. These advances achieve superior computational efficiency, fidelity, and improved protein expression compared to earlier methods. Comparative analyses have also highlighted the importance of integrating multiple metrics in the objective, such as CAI, GC content, mRNA folding energy, and codon pair bias, rather than relying on single-metric approaches~\citep{zhang2023lineardesign}.

\paragraph{Discrete Sequence Optimization}
Methods for optimizing discrete sequences vary along two axes. The first is the \emph{search space geometry}: whether the optimizer works directly in sequence space, or in a learned continuous latent space that is decoded back into sequences. The second is the \emph{trust mechanism}: what stops the optimizer from exploiting the surrogate model where it is unreliable, i.e., on out-of-distribution designs (e.g., via a conservative surrogate, posterior uncertainty, etc.). We organize prior work by search space geometry, and comment on the trust mechanism in each case.

\paragraph{Sequence-Space Search}
The first geometry operates directly on sequences (or distributions over them). Population-based methods evolve candidates through mutation and selection: Genetic Algorithms (GA)~\citep{deb2002fast} are simple but sample-inefficient when evaluations are costly~\citep{turner2021bayesian}, and model-based variants add a learned model to screen proposals~\citep{yang2019machine} or guide mutation~\citep{zhang2021unifying}. Estimation-of-distribution methods instead maintain a learned distribution $q_\theta(x)$ over sequences, repeatedly sampling, reweighting toward high fitness, and refitting~\citep{brookes2019dbas, brookes2019conditioning}. The VAE in \citep{brookes2019conditioning} acts only as a density \emph{prior} to keep sampling on the same manifold, so the search still lives in sequence space. A second line of work trains a fitness surrogate and runs gradient ascent on a continuous relaxation of the sequence. This approach can fail miserably as the surrogate tends to over-predict on the very inputs the optimizer discovers, since they drift off the training distribution: a failure mode known as reward hacking~\citep{trabucco2021conservative, trabucco2022design, kim2025offline}. The proposed remedies adjust the \emph{trust mechanism} rather than the geometry: conservative objective models regularize the surrogate to under-predict off-distribution~\citep{trabucco2021conservative}; Reinforcement Learning methods generate tokens under a surrogate reward with per-round model selection~\citep{angermueller2020dynappo}; and Bayesian optimization over string kernels or embeddings uses posterior uncertainty and trust regions~\citep{benjamins2024bayesian}.

\paragraph{Latent-Space Optimization}
The second geometry learns a continuous latent space, shared by a decoder that reconstructs sequences and a model that predicts their properties, and optimizes in that latent space~\citep{gomez2018automatic, jin2021iterative, tripp2020sample}. Latent-Space Bayesian Optimization (LSBO) adds a probabilistic surrogate, generally a Gaussian process or Deep Ensemble, to quantify epistemic uncertainty and balance exploration-exploitation across rounds of propose, test, and update~\citep{stanton2022accelerating, gruver2023protein, amin2024bayesian}. The upside is that a smooth, low-dimensional space makes gradient-based and Bayesian search easy to apply and additionally acts as a prior that keeps candidates realistic. The downside is the extra decoder, which needs to be trained and can potentially mis-decode, creating the chance for a design to fall off-distribution. `Reward hacking' is therefore not unique to sequence-space search, and motivates the use of uncertainty safeguards.

\paragraph{Our approach}
LSCO belongs to the latent-space family but differs in three ways. First, we optimize for \emph{predicted expression} directly, not proxy metrics, within a multi-criteria objective that also weighs mRNA stability and naturalness while maintaining target protein fidelity. Second, rather than training a task-specific autoencoder, we pre-train and reuse an mRNA foundation model and conduct search in its latent space, inheriting realistic codon-usage priors. Third, our expression surrogate model is trained as a deep ensemble, whose model/epistemic uncertainty steers the search away from unreliable regions.

\section{Implementation details}
\label{appsec: implementation}

\paragraph{Back-translation model}
The back-translation model uses the standard BERT architecture with absolute position embeddings, configured with a hidden dimension of 777, 9 transformer layers, 21 attention heads per layer, a feed-forward dimension of 777, and a maximum sequence length of 2048 tokens. It operates over a joint vocabulary of 91 tokens: 27 amino-acid tokens (20 canonical amino acids plus 7 special tokens \texttt{[BOS]}, \texttt{[EOS]}, \texttt{[CLS]}, \texttt{[SEP]}, \texttt{[MASK]}, \texttt{[PAD]}, and \texttt{[UNK]}) as the input alphabet, and 64 codon tokens (one per sense codon) as the output alphabet. During training, amino-acid input tokens are masked with probability $p = 0.2$.

\paragraph{Variational mRNA language model}
We use a 12-layer ModernBERT architecture \citep{modernbert} with Rotary Position Embeddings (RoPE) \citep{su2024roformer}. The model uses a vocabulary of 70 tokens (64 codons plus 6 special tokens \texttt{[BOS]}, \texttt{[EOS]}, \texttt{[CLS]}, \texttt{[SEP]}, \texttt{[MASK]}, and \texttt{[PAD]}), a hidden dimension of 768, 12 attention heads per layer, and a feed-forward dimension of 3072. The VMAE masking ratio is set to \(p=0.15\). The model is split at layer \(k=6\), using the first six layers as the encoder \(E\) and the remaining six as the decoder \(D\). We use a standard Gaussian prior \(\mathcal{N}(0,I)\) over the latent variables \(\mathbf{z}\in\mathcal{Z}_L\). Pretraining is performed for 40 epochs using AdamW \citep{loshchilov2017decoupled} with a batch size of 256, an initial learning rate of \(10^{-4}\), linear warmup over the first 10\% of training steps, and cosine learning-rate decay thereafter. 

\paragraph{Expression predictor}
We implement \(\Psi\) as a dual-chain transformer that models heavy and light chain mRNA separately, then aggregates the two chain representations using Multiple Instance Learning (MIL) attention before a final linear layer predicts antibody expression. In practice, we found that reusing the pretrained language model encoder \(E\) as the backbone for \(\Psi\) improves the expression prediction. To enable this, we replace the first \texttt{nn.Embedding} layer of \(E\) with a dense \texttt{nn.Linear} layer to handle soft non-one-hot sequences. Once trained, \(\Psi\) is differentiable with respect to the soft-sequence inputs \(\tilde{\mathbf{r}}_{\tau}\), allowing gradient-based optimization. Training proceeds for 120 epochs with Negative Log-Likelihood loss (for each member $m$ of the ensemble) on the mAb expression (absorbance measured at 280 nm) data \citep{prakash2024bridging,yazdani2025helm}. We use AdamW optimization with a batch size of 64, initial learning rate of \(5 \times 10^{-5}\), and the same warmup and decay schedule as the language model. The Spearman's rank correlation between the predicted and ground truth expression on the test split reaches $0.65$.

We operate $\Psi$ on the relaxed sequence $\tilde{\mathbf{r}}_{\tau}$ rather than the latent space $\mathcal{Z}_L$ for two reasons: (i) early experiments showed that optimizing directly in $\mathcal{Z}_L$ can cause large shifts in predicted expression and latent values without corresponding changes at the sequence level; and (ii) the softmax representation admits the simple constrained-decoding step used to enforce $\tilde{\mathbf{r}}_{\tau} \in \mathcal{S}(\mathbf{a})$.

\begin{algorithm}[t]
\caption{Latent-Space Codon Optimizer (LSCO)}
\label{alg:lsco}
\begin{algorithmic}[1]
\REQUIRE Protein sequence \(\mathbf{a}\), pretrained encoder \(E\), decoder \(D\), predictors \(\Psi\) and \(\Upsilon\), weights \(\lambda_{\Psi}, \lambda_{\Upsilon}, \lambda_{\text{nat}}\), initial \(\tau_{\text{high}}\), final \(\tau_{\text{low}}\), optimization steps \(T\), learning rate \(\eta\)
\ENSURE Optimized mRNA sequence \(\mathbf{c}^\star\)
\STATE Initialize \(\mathbf{z} \in \mathcal{Z}_L\) \COMMENT{randomly from \(\mathcal{N}(0,I)\)}
\FOR{\(t = 1\) to \(T\)}
    \STATE \(\tau \gets \tau_{\text{high}} + (t/T)\,(\tau_{\text{low}} - \tau_{\text{high}})\) \COMMENT{temperature annealing}
    \STATE \(\tilde{\mathbf{r}}_{\tau} \gets \softmax_{\tau}(D(\mathbf{z}) + M(\mathbf{a}))\) \COMMENT{constrained soft sequence}
    \STATE \(\mathcal{L} \gets \lambda_{\Psi} U_{\Psi}(\tilde{\mathbf{r}}_{\tau}) - \lambda_\text{MFE}\, \Upsilon(\tilde{\mathbf{r}}_{\tau}) + \lambda_{\text{nat}}\, U_{\text{nat}}(\mathbf{z}, \mathbf{z}_{\text{nat}})\)
    \STATE \(\mathbf{z} \gets \mathbf{z} + \eta\, \nabla_{\mathbf{z}} \mathcal{L}\) \COMMENT{gradient \emph{ascent} on the objective}
\ENDFOR
\STATE \(\mathbf{r} \gets \argmax(D(\mathbf{z}) + M(\mathbf{a}))\) \COMMENT{discrete one-hot mRNA}
\STATE \(\mathbf{c}^\star \gets\) map \(\mathbf{r}\) to codon sequence
\RETURN \(\mathbf{c}^{\star}\)
\end{algorithmic}
\end{algorithm}

\paragraph{MFE predictor}
Analogously, the MFE predictor \(\Upsilon\) mirrors \(\Psi\)'s architecture. It is trained for 50 epochs using MSE loss on 50,000 sub-sampled OAS mRNA sequences, with ground-truth MFE labels computed using the ViennaRNA package. Optimization follows identical settings to \(\Psi\). The Spearman's rank correlation between the predicted and ground truth MFE on the test split reaches $0.91$.

\paragraph{Constrained Decoding}
Both the back-translation model and the optimizer require that every decoded sequence translates exactly to the target protein $\mathbf{a}$, enforced with a shared masking mechanism. Let $\mathbf{l} = D'(\mathbf{z}) \in \mathbb{R}^{L\times64}$ be the pre-softmax decoder logits. For each position $a_j$ we define a validity mask
\begin{equation}
    M(a_j)[i] =
    \begin{cases}
        0, & \text{if codon } i \in \mathcal{C}(a_j), \\
        -\infty, & \text{otherwise},
    \end{cases}
\end{equation}
and compute masked logits $l_{j,i} \leftarrow l_{j,i} + M(a_j)[i]$, then apply the temperature-$\tau$ softmax $\tilde{\mathbf{r}}_{\tau} = \softmax_{\tau}(\mathbf{l})$. This guarantees $\tilde{\mathbf{r}}_{\tau} \in \mathcal{S}(\mathbf{a})$ while allowing free exploration of synonymous codons at each position.

\paragraph{LSCO hyperparameters}
We determined the weights \(\lambda_{\Psi}\) and \(\lambda_{\Upsilon}\) via grid search on each dataset to maximize predicted A280 while ensuring the optimized sequence's MFE is generally lower than the wild-type/original MFE, with flexibility to relax this constraint when predicted A280 $\leq 1$. In practice, LSCO proved robust to a wide range of these weights. Grid search identified \(\lambda_{\Psi} = 10\) and \(\lambda_{\Upsilon} = 0.1\) as effective. For decoding, we set the initial temperature \(\tau_{\text{high}} = 3\) and the final temperature \(\tau_{\text{low}} = 0.01\). These temperature parameters showed higher sensitivity, particularly \(\tau_{\text{low}}\); excessively high values caused significant rounding errors, reducing expression, while overly low values hindered optimization progress. We ran LSCO for \(T = 3000\) iterations, observing that additional iterations improved convergence marginally. The learning rate was set to \(\eta = 10^{-4}\), and, leveraging the differentiability of the LSCO objective, we employed the Adam optimizer for robust gradient-based updates.

\paragraph{Decoding Temperature} In practice, we also observed the importance of carefully selecting the decoding temperature \(\tau\). Too high temperature allows for broader exploration of the synonymous sequence space but can introduce significant rounding errors when mapping \(\Tilde{\mathbf{r}}_{\tau}\) to the nearest discrete mRNA sequence \(\mathbf{r}\) via argmax. Conversely, too low temperature hinders exploration of the soft-sequence space by creating high-energy barriers between sequences. Empirically, we found that linearly annealing \(\tau\) from a high to a low value during optimization strikes the best balance. This strategy allows the optimizer to first explore the domain freely to identify promising candidates and then commit to a single solution at the end of the process

\section{Additional Results and Experiments}


\begin{figure}[t]
    \centering
    \includegraphics[width=\linewidth]{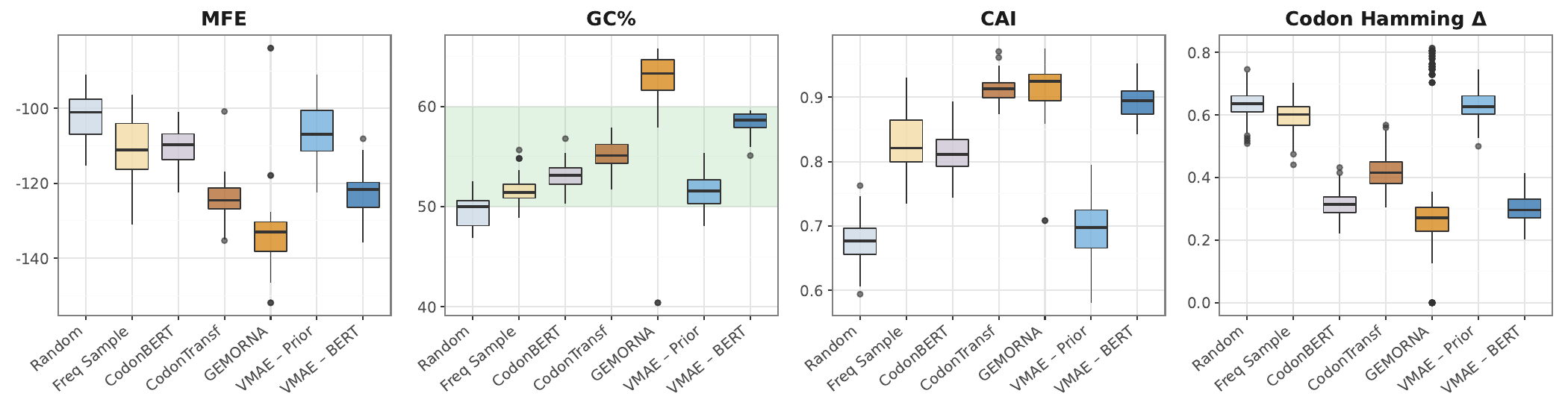}
    \caption{\textbf{Sequence-level biophysical proxies across generation methods.}
    Minimum Free Energy (MFE; lower = more stable), GC Content (GC\%; shaded band marks host-appropriate range), Codon Adaptation Index (CAI; higher is better), and codon Hamming distance ($\Delta$) from the reference sequence (lower = fewer codon edits), for non-contextual baselines (\textit{Random}, \textit{Freq Sample}), modern learned generators (\textit{CodonBERT}, \textit{CodonTransf}, \textit{GEMORNA}), and our pretrained models (\textit{VMAE--Prior}, decoding from Gaussian latent prior; \textit{VMAE--BERT}, decoding from CodonBERT generated sequence's latent space). 
    }
    \label{fig:pretrained_model_metrics}
\end{figure}

\subsection{Pretrained mRNA language models comparison}

To evaluate whether pretrained mRNA language models learn biologically meaningful codon patterns, we compared four sequence generation strategies: random codon assignment (\textit{Random}), codon sampling according to empirical codon frequencies (\textit{Freq Sample}), an amino-acid-to-codon translation model (\textit{CodonBERT})
pretrained on OAS sequences, and a codon-to-codon Variational Masked Auto-Encoder (\textit{VMAE}) also pretrained on the OAS data. For VMAE generation, we include two variants, one that samples new sequences starting from the prior latent space (\textit{VMAE – Prior}) and one that samples sequences starting from the encoded CodonBERT generated sequences (\textit{VMAE – BERT}). All sequences are evaluated on Minimum Free Energy (MFE), GC content, Codon Adaptation Index (CAI), and codon-level Hamming distance from the reference starting sequence.

As shown in Figure~\ref{fig:pretrained_model_metrics}, modern pretrained mRNA language models generate sequences with substantially improved biological properties than other baselines. CodonTransf, GEMORNA and VMAE – BERT all score favorably in terms of lower MFE, higher GC content (with GEMORNA falling just above the ideal host-specific GC range), higher CAI values, and lower codon Hamming distances. This indicates that modern transformer-based pretrained models efficiently learn context-dependent synonymous codon substitutions that improve quality without unnecessary sequence changes.

\subsection{LSCO Optimization Trajectory}
\label{app:latent_traj}

\begin{figure}[t]
    \centering
    \includegraphics[width=0.95\linewidth]{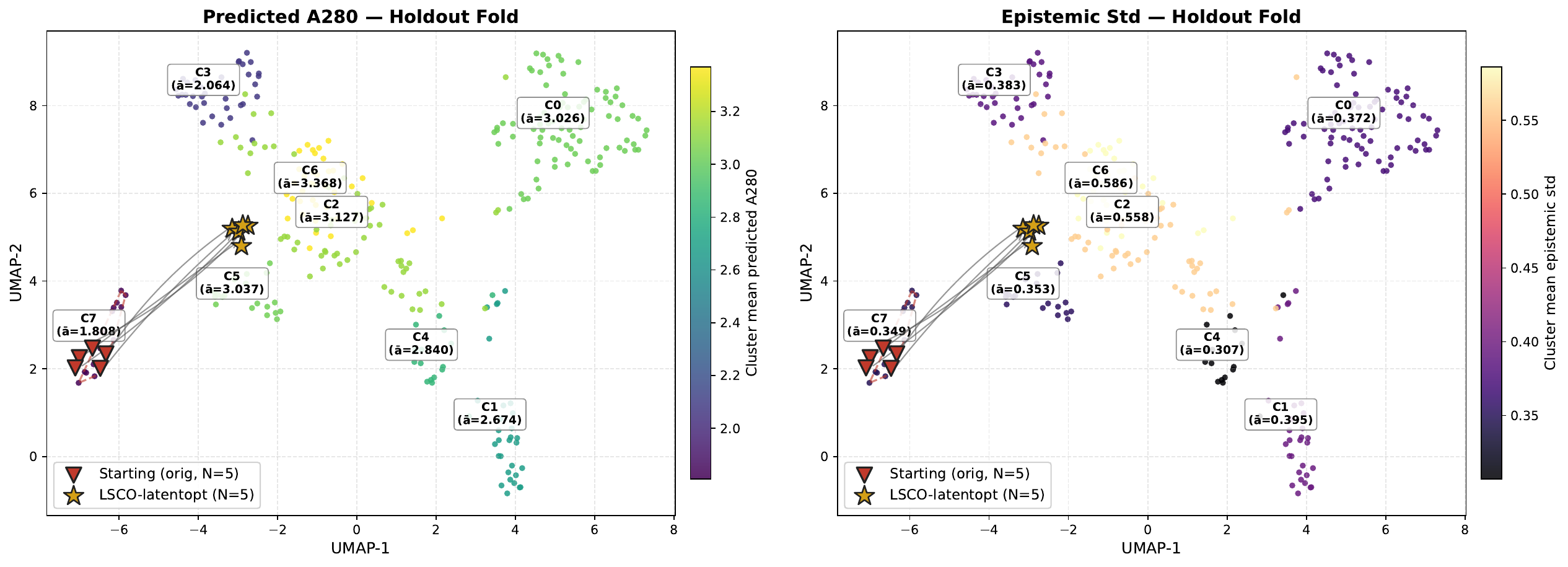}
    \caption{\textbf{LSCO optimization trajectories in latent space.} UMAP projection of the pretrained VMAE latent space on a holdout fold, coloured by predicted A280 (left) and by predictor epistemic uncertainty (right). Starting sequences (red triangles) lie in a low-expression cluster (C7, $\bar{a}=1.808$); LSCO-optimized sequences (gold stars) are relocated toward high-expression clusters ($\bar{a}\approx 3.0$--$3.4$). Trajectories (grey arrows) follow a consistent direction and remain in low-uncertainty regions.}
    \label{fig:UMAP_traj}
\end{figure}

Figure~\ref{fig:UMAP_traj} illustrates how LSCO reshapes the distribution of sequences in the learned latent space. The starting sequences (red triangles) reside in a cluster (C7) with low averaged predicted expression ($\bar{a}=1.808$), whereas the LSCO-optimized sequences (gold stars) are consistently relocated toward clusters with substantially higher averaged predicted A280 ($\bar{a}\approx 3.0$--$3.4$). All optimization trajectories follow a similar direction, indicating that LSCO discovers coherent ascent paths on the underlying fitness landscape rather than arbitrary sequence perturbations. The companion uncertainty panel (right) shows that the optimized sequences stay in regions of low epistemic uncertainty, confirming that the LCB utility keeps the search in-distribution rather than exploiting the predictor off-manifold.

\subsection{Codon-Usage Analysis}
\label{app:codon_analysis}

\begin{figure}[h]
    \centering
    \includegraphics[width=0.9\linewidth]{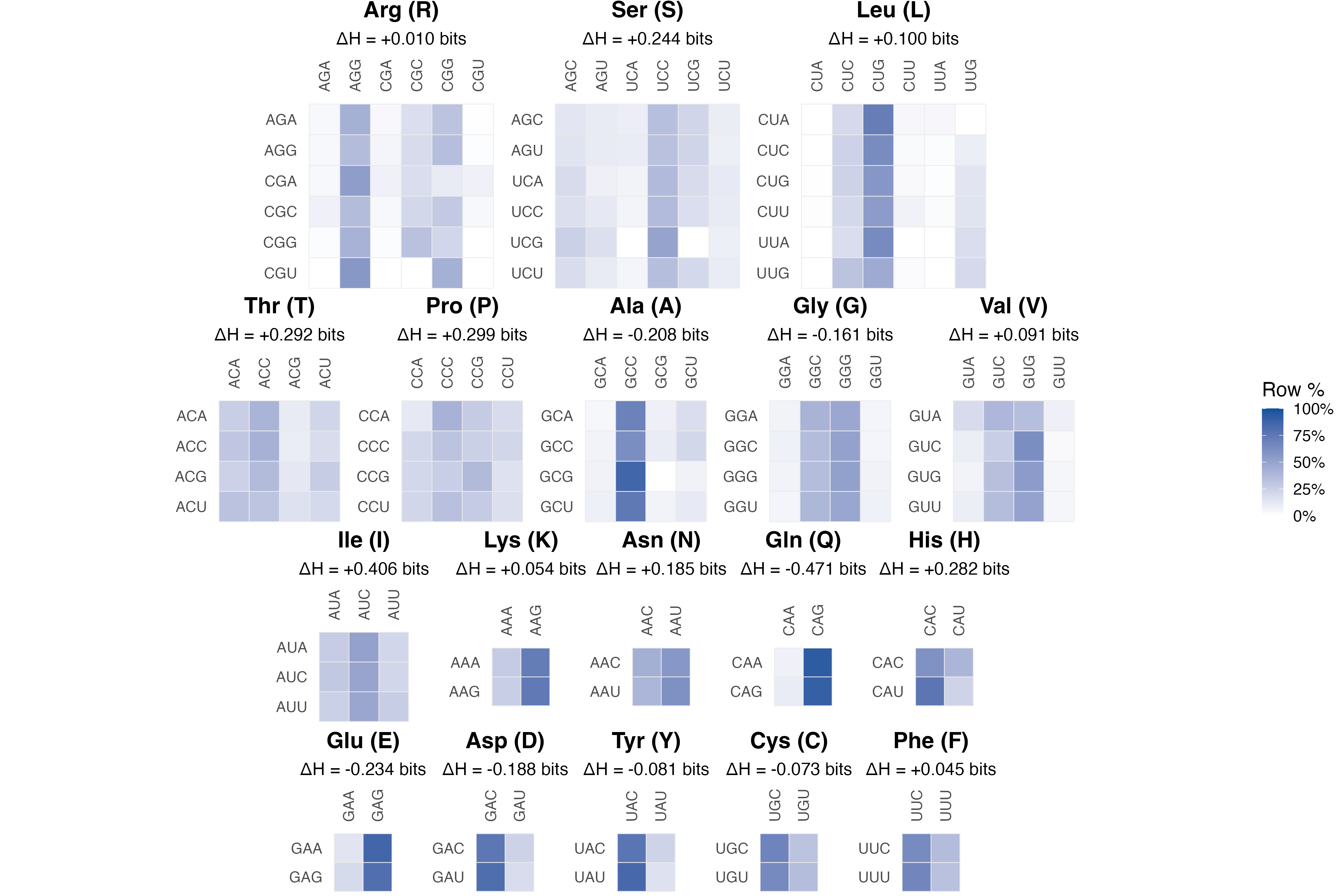}
    \caption{\textbf{Codon substitutions induced by LSCO optimization.} Rows denote original codons, columns optimized codons; colours show row-normalized transition probabilities for synonymous substitutions produced by \textit{LSCO -- LCB + MFE}, over the 54 starting mAbs. Strong diagonal structure indicates codon conservation, while off-diagonal transitions reveal learned preferences. The Symmetrized KL (SKL) divergence quantifies the magnitude of codon redistribution per amino-acid family (higher SKL $\rightarrow$ larger redistribution).}
    \label{fig:CU_heatmap}
\end{figure}

To interpret the strategy learned by \textit{LSCO -- LCB + MFE}, Figure~\ref{fig:CU_heatmap} shows codon-level transition probabilities between original and optimized sequences, broken down by amino acid, over the 54 selected mAbs, i.e.\ which synonymous codons LSCO tends to replace and which it preserves. Several amino-acid families exhibit strong off-diagonal structure, indicating that optimization consistently redirects codon usage toward a preferred subset of alternatives rather than making random synonymous substitutions. There is also a moderate shift toward GC-richer codons across multiple families, consistent with the GC-content and MFE improvements of \textit{LSCO -- LCB + MFE} in Figure~\ref{fig:main_results}. However, the transition patterns are highly amino-acid-specific, suggesting a more nuanced strategy than simple GC maximization; the Symmetrized KL values confirm that the magnitude of redistribution varies substantially across families, as LSCO concentrates its modifications on specific synonymous groups.

\subsection{A280 Predictors and Agreement}
\label{app:agreement}

\paragraph{1D Sequence-Based Predictors.}
Both 1D predictors follow the same dual-chain design: heavy and light chain mRNAs are encoded separately, each chain is reduced to a fixed-size representation, and the two are aggregated with a Multiple Instance Learning (MIL) attention module before a final linear layer predicts A280. The two variants then differ mainly in the encoder backbone. The \emph{native} predictor ($\Psi$) reuses the pretrained VMAE encoder $E$ as its backbone (with the first embedding layer replaced by a dense layer so it accepts the relaxed, soft-sequence input $\tilde{\mathbf{r}}_\tau$); this is the model whose gradients drive LSCO. The \emph{external} 1D predictor keeps similar dual-chain + MIL head but swaps in a different, shallower autoencoder backbone that never takes part in optimization. It therefore provides an independent read-out that shares the 1D sequence view but not the representation LSCO is optimized against.

\paragraph{2D Structure-Based Predictor.}
The external ``2D GNN'' predictor takes a genuinely different view of each sequence. For each chain, the mRNA is folded and its dot-bracket secondary structure is converted into a graph, with one-hot nucleotide node features and edges from both the backbone and the predicted base pairs. A graph neural network encodes this structure, and the same MIL-attention aggregation combines the two chains before the regression head. Because it consumes secondary-structure graphs rather than raw sequence embeddings, the 2D GNN carries a structural inductive bias absent from both 1D models, making it the most independent of the three evaluators.

\begin{figure}
    \centering
    \begin{subfigure}{0.49\textwidth}
    \centering
    \includegraphics[width=0.95\linewidth]{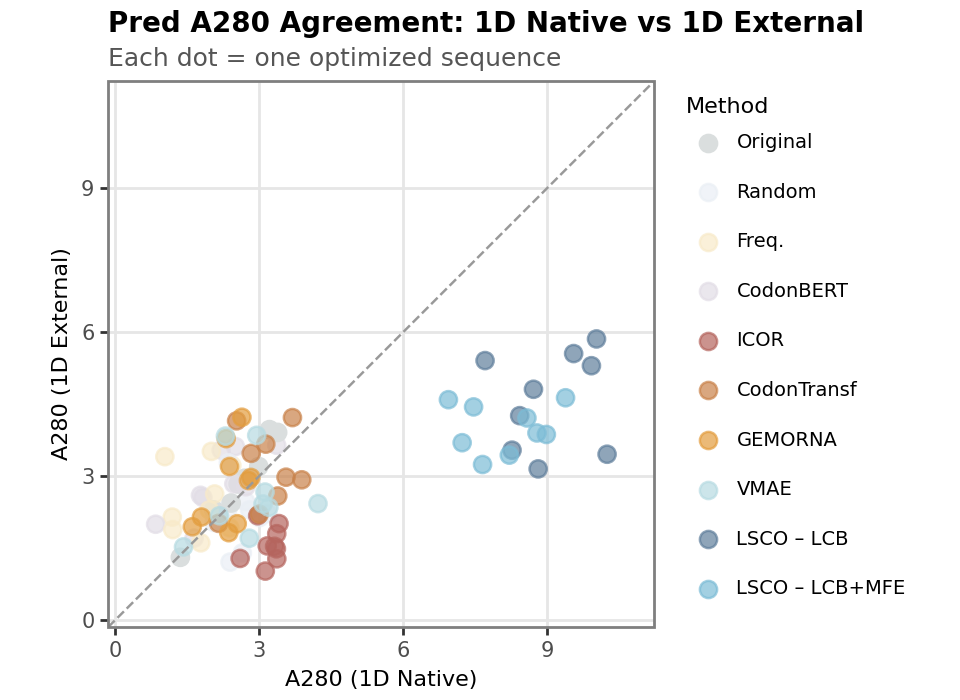}
    \caption{Native 1D vs External 1D}
    \label{fig:agree_1d}\end{subfigure}
\hfill
\begin{subfigure}{0.49\textwidth}
    \centering
    \includegraphics[width=0.95\linewidth]{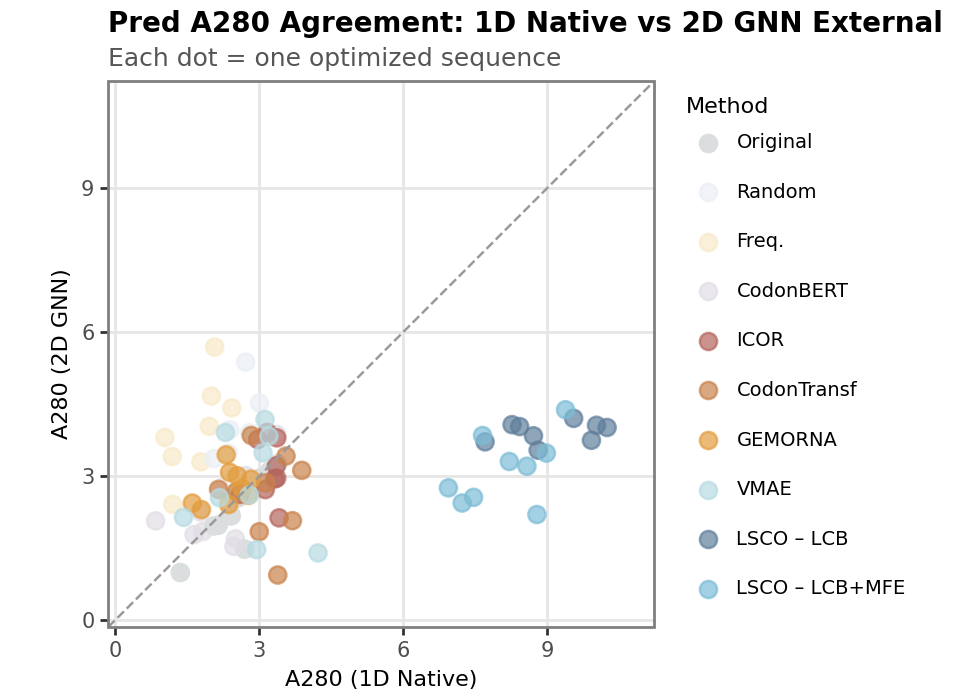}
    \caption{Native 1D vs External 2D GNN.}
    \label{fig:agree_2d}
\end{subfigure}
\caption{\textbf{Predicted-A280 agreement across evaluators.} Each dot is one optimized sequence, coloured by method; the dashed line marks perfect agreement. The horizontal axis is the native 1D predictor $\Psi$; the vertical axis is an independent external predictor withheld from optimization, the 1D Shallow-AE model (a) and the structure-aware 2D GNN (b). LSCO variants (blue) are rated highest by the native model, but score favorably also under the two external models in terms of ordering/ranking preserving.}
\label{fig:combined_a280_agree}
\end{figure}

\paragraph{Cross-predictor agreement.}
Figure~\ref{fig:combined_a280_agree} plots, for every optimized sequence, the native 1D A280 against each external predictor. In both panels the LSCO points (blue) sit clearly at the center-right, separated from all baselines: this implies they are rated highest by the native predictor and rated comparably or slightly higher by the external 1D and the structure-aware 2D GNN. The external predictors are more conservative in absolute terms in that most points lie below the diagonal, so agreement is more on \emph{ranking} rather than exact value prediction. Overall ordering is more or less preserved across all three, despite distinct structural views. This consistency across predictors with different backbones and inductive biases presents some evidence that the gains are not an artifact of $\Psi$ being exploited off-distribution, but reflect genuinely higher predicted expression.

\subsection{Deep-Ensemble and Epistemic Uncertainty Aware Objective}
\label{app:deepens}

\paragraph{Probabilistic Deep Ensemble.}
We implement the expression predictor $\Psi$ as a deep ensemble of $M=10$ \emph{probabilistic} members~\citep{lakshminarayanan2017simple}. Each member $m$ does not output a point estimate but a full Gaussian predictive distribution over A280:
\begin{equation}
p_m\!\left(y \mid \tilde{\mathbf{r}}_\tau\right) = \mathcal{N}\!\left(y;\, \mu_m(\tilde{\mathbf{r}}_\tau),\, \sigma_m^2(\tilde{\mathbf{r}}_\tau)\right),
\end{equation}
where the head outputs both a mean $\mu_m$ and a variance $\sigma_m^2$. Members are trained \emph{independently}, each with its own random initialization, optimizer, and backward pass, by minimizing the Gaussian negative log-likelihood loss:
\begin{equation}
\mathcal{L}_{\text{NLL}}^{(m)} = \tfrac{1}{2}\log \sigma_m^2(\tilde{\mathbf{r}}_\tau) + \frac{\left(y - \mu_m(\tilde{\mathbf{r}}_\tau)\right)^2}{2\,\sigma_m^2(\tilde{\mathbf{r}}_\tau)},
\end{equation}
so that the predicted variance is calibrated to each member's own residuals. The variance head itself is configurable (constant, scalar, per-group, or fully heteroscedastic); we use the scalar $\sigma_m^2$ setting, as we noticed data exhibit somehow uniform noise.

\paragraph{Aleatoric/Epistemic Uncertainty Decomposition.}
Treating the ensemble as a uniformly-weighted Gaussian mixture, the ensemble mean and total predictive variance are
\begin{align}
\mu_\Psi(\tilde{\mathbf{r}}_\tau) &= \frac{1}{M}\sum_{m=1}^{M}\mu_m(\tilde{\mathbf{r}}_\tau), \\
\sigma_\Psi^2(\tilde{\mathbf{r}}_\tau) &= \underbrace{\frac{1}{M}\sum_{m=1}^{M}\sigma_m^2(\tilde{\mathbf{r}}_\tau)}_{\text{aleatoric}} + \underbrace{\frac{1}{M}\sum_{m=1}^{M}\big(\mu_m(\tilde{\mathbf{r}}_\tau) - \mu_\Psi(\tilde{\mathbf{r}}_\tau)\big)^2}_{\text{epistemic}}.
\end{align}
The two terms have distinct meanings. The \emph{aleatoric} term is the average of the members' own predicted variances and captures irreducible noise in the measurement/data. The \emph{epistemic} term is the variance \emph{across} the members' means and captures model uncertainty: it is large where the members disagree, which happens precisely in regions of sequence space that are under-represented in the training data. Because epistemic uncertainty shrinks as designs stay close to the training distribution and grows off-distribution, it is exactly the signal the LCB utility $U_\Psi = \mu_\Psi - \lambda_{\text{LCB}}\,\sigma_\Psi$ (Eq.~\ref{eq:expression_utility}) exploits: penalizing $\sigma_\Psi$ steers the optimizer away from over-optimistic, out-of-distribution sequences. The same probabilistic ensemble construction is used for all three predictors (1D native, 1D external, 2D GNN); the 2D GNN members simply own an independent structure-graph backbone each.

\begin{table}
    \centering
    \resizebox{\linewidth}{!}{%
    \begin{tabular}{cccccccc}
        \toprule
         & & \multicolumn{3}{c}{Predicted A280} & \multicolumn{3}{c}{Metrics (VH + VL)} \\
        \cmidrule(lr){3-5} \cmidrule(lr){6-8}
        $\lambda_{MFE}$ & $\lambda_{LCB}$ & 1D Transf (Internal) & 1D Transf (External) & 2D GNN & MFE & GC\% & CAI \\
        \midrule
        \multirow{2}{*}{0}   & 0 & $\mathbf{9.585 \pm 0.218}$ & $\mathbf{4.518 \pm 0.223}$ & $\mathbf{3.492 \pm 0.203}$ & $-235.187 \pm 3.555$         & $53.303 \pm 0.423$ & $0.816 \pm 0.008$ \\
                             & 1 & $9.156 \pm 0.241$          & $4.400 \pm 0.254$          & $3.485 \pm 0.212$          & $-236.711 \pm 3.603$         & $53.601 \pm 0.390$ & $0.815 \pm 0.009$ \\
        \midrule
        \multirow{2}{*}{0.1} & 0 & $8.180 \pm 0.237$ & $3.900 \pm 0.224$ & $3.175 \pm 0.217$ & $-266.274 \pm 2.783$ & $57.756 \pm 0.285$ & $0.830 \pm 0.009$ \\
                             & 1 & $7.858 \pm 0.235$ & $3.935 \pm 0.265$ & $3.099 \pm 0.187$ & $-267.050 \pm 3.211$ & $57.916 \pm 0.300$ & $0.837 \pm 0.008$ \\
        \midrule
        \multirow{2}{*}{1}   & 0 & $8.065 \pm 0.232$ & $3.887 \pm 0.200$ & $3.319 \pm 0.201$ & $-263.765 \pm 3.122$ & $58.003 \pm 0.304$ & $0.825 \pm 0.009$ \\
                             & 1 & $7.899 \pm 0.256$ & $3.951 \pm 0.271$ & $2.923 \pm 0.182$ & $-266.365 \pm 3.381$ & $58.018 \pm 0.317$ & $\mathbf{0.838 \pm 0.008}$ \\
        \midrule
        \multirow{2}{*}{5}   & 0 & $7.921 \pm 0.260$ & $3.729 \pm 0.220$ & $3.057 \pm 0.220$ & $-265.006 \pm 3.443$ & $57.959 \pm 0.307$ & $0.824 \pm 0.009$ \\
                             & 1 & $7.635 \pm 0.271$ & $3.852 \pm 0.228$ & $3.161 \pm 0.249$ & $-263.954 \pm 3.377$ & $57.973 \pm 0.321$ & $0.832 \pm 0.008$ \\
        \midrule
        \multirow{2}{*}{10}  & 0 & $7.490 \pm 0.248$ & $3.514 \pm 0.263$ & $3.089 \pm 0.248$ & $\mathbf{-267.381 \pm 3.033}$ & $58.494 \pm 0.282$ & $0.818 \pm 0.008$ \\
                             & 1 & $7.390 \pm 0.264$ & $3.500 \pm 0.256$ & $3.169 \pm 0.241$ & $-265.220 \pm 2.735$         & $58.233 \pm 0.258$ & $0.829 \pm 0.009$ \\
    \end{tabular}%
    }
    \caption{\textbf{MFE sensitivity analysis for LSCO variants.} Predicted A280 values from three surrogate models (1D Transformer (Internal), 1D Transformer (External), 2D GNN) and biophysical metrics (minimum free energy, GC content, codon adaptation index) reported for the joint heavy + light chain construct on the 54 starting sequences. The second row ($\lambda_{MFE}=0, \lambda_{LCB}=1$) is the LCB-only LSCO baseline (`LSCO -- LCB' in Figure \ref{fig:main_results}). The fourth row ($\lambda_{MFE}=0.1, \lambda_{LCB}=1$) is the LCB+MFE baseline (`LSCO -- LCB + MFE' in Figure \ref{fig:main_results}) that best balances A280 and MFE objectives. \textbf{Bold} indicates the best value per column: highest for A280 predictions and CAI, most negative for MFE. GC\% is not bolded as ``best'' is target-dependent. Values are mean $\pm$ 95\% confidence bands across starting sequences.}
    \label{tab:mfe_sensitivity_ver2}
\end{table}

\begin{figure}
    \centering
    \includegraphics[width=\linewidth]{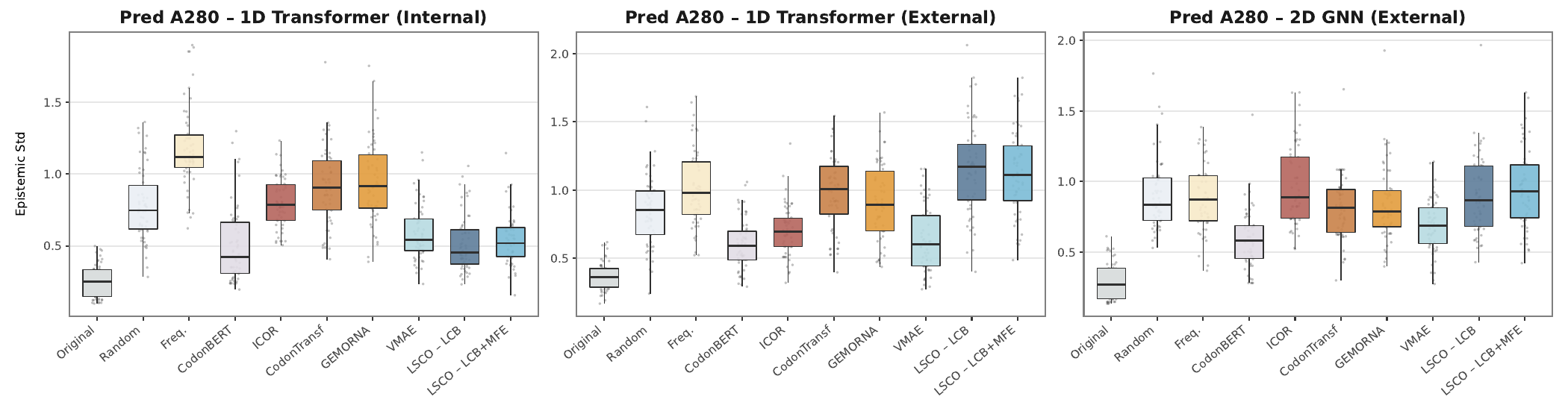}
    \caption{\textbf{Epistemic uncertainty of the optimized sequences.} Epistemic standard deviation (across-member disagreement, $\sqrt{\text{epistemic}}$) of the three probabilistic deep-ensemble predictors — 1D Transformer Internal/Native (left), 1D Transformer External (middle), 2D GNN External (right) — evaluated on the same 9 optimized mAbs as Figure~\ref{fig:main_results}; each dot is one mAb. Under the native predictor, LSCO variants (blue) show low, concentrated epistemic uncertainty, comparable to the unoptimized VMAE and well below the other optimizers, confirming the LCB utility keeps the search in-distribution. Under the external predictors — never seen during optimization — LSCO's uncertainty is comparable to other high-expression methods, as expected for held-out evaluators.}
    \label{fig:epist_std}
\end{figure}

\paragraph{Epistemic uncertainty of the optimized sequences.}
Figure~\ref{fig:epist_std} reports the epistemic standard deviation ($\sqrt{\text{epistemic}}$) of each predictor on the same 9 optimized mAbs as Figure~\ref{fig:main_results}. Two patterns stand out. First, under the \emph{native} predictor (left), the LSCO variants sit at low and tightly-concentrated epistemic uncertainty, comparable to the unoptimized VMAE sequences and well below frequency-based, ICOR, CodonTransformer, and GEMORNA. This confirms that the LCB penalty keeps the search in-distribution rather than chasing high-variance designs. Second, under the two \emph{external} predictors (middle, right), LSCO's epistemic uncertainty rises to a level similar to the other high-expression methods. This is because the external models were never optimized against, so any strong-expression design is somewhat novel to them, and LSCO is no more off-distribution for them than the competing optimizers. Read together with Figure~\ref{fig:main_results}, LSCO therefore attains the \emph{highest} predicted A280 across all three predictors while keeping the \emph{lowest} epistemic uncertainty under the predictor that actually guides optimization, i.e., its gains do not come from exploiting high-uncertainty regions of its own surrogate.

\subsection{MFE regularization sensitivity}
\label{app:mfe_sens}

Table~\ref{tab:mfe_sensitivity_ver2} reports the full sweep over the MFE weight $\lambda_{\text{MFE}}$ and the uncertainty-penalty toggle $\lambda_{\text{LCB}}$, extending the two settings shown in the main text (Figure~\ref{fig:main_results}). Increasing $\lambda_{\text{MFE}}$ from $0$ drives MFE substantially lower and lifts GC and CAI into the host-appropriate range, at a modest cost in predicted A280; gains in stability saturate beyond $\lambda_{\text{MFE}}=0.1$, confirming it as the sweet spot used throughout.


\subsection{Seed sensitivity study}
\label{app:seed_study}

\begin{figure}[t]
    \centering
    \includegraphics[width=\linewidth]{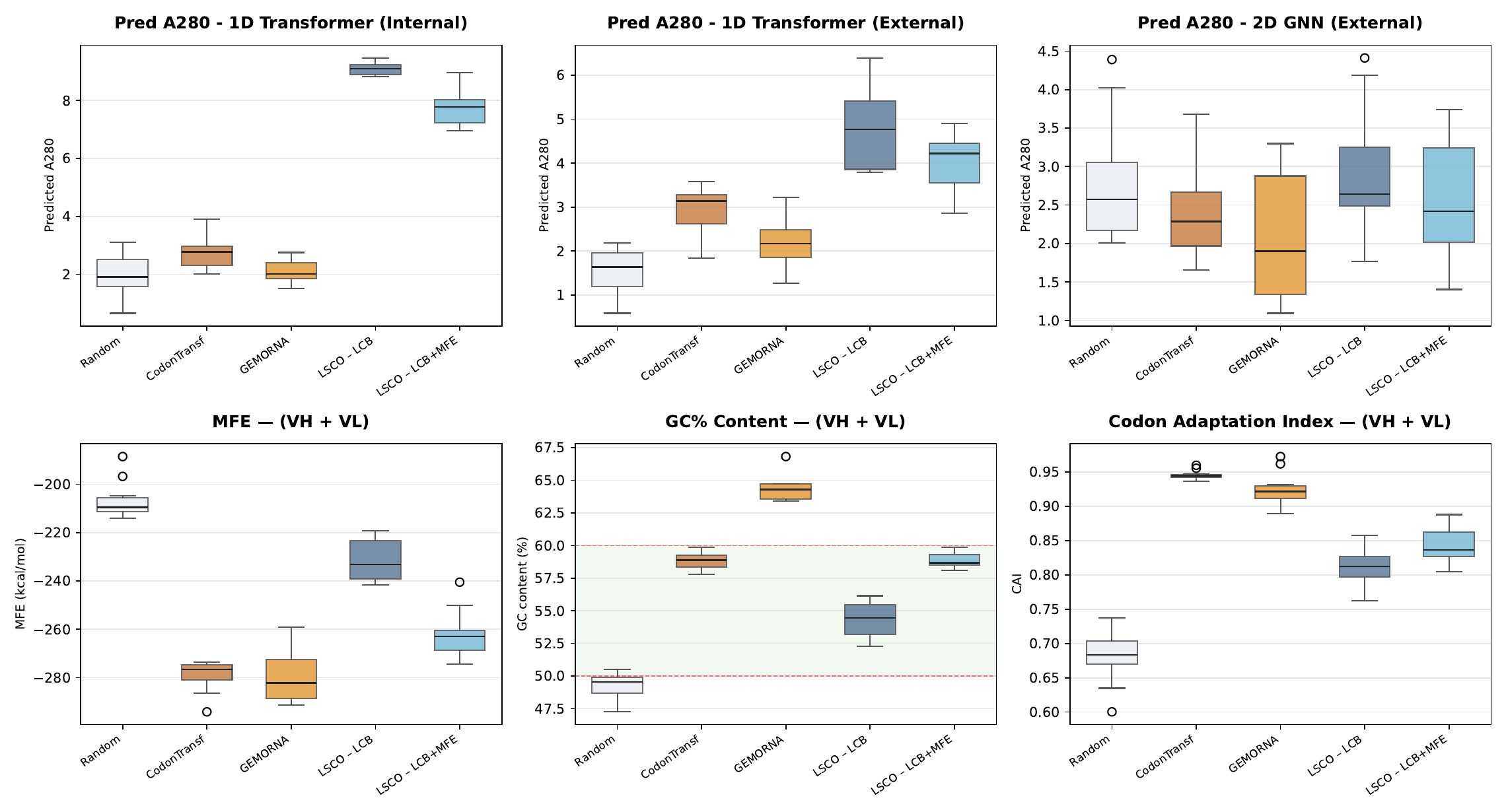}
    \caption{\textbf{Seed sensitivity on a single mAb (10 seeds).} Boxes show the spread over different 10 seeded runs, reflecting the stochasticity behind different codon optimization methods. \emph{Top:} predicted A280 from the native 1D transformer (internal), an external 1D transformer, and an external 2D GNN. \emph{Bottom:} biophysical metrics for the joint heavy+light construct (VH+VL): MFE (lower = more stable), GC\% (shaded band = host-appropriate $50$--$60\%$), and CAI. Both LSCO variants remain above the stochastic baselines on the 1D predictors across all seeds with bounded spread, keep GC within the admissible band, and \textit{LSCO -- LCB + MFE} consistently lowers MFE relative to \textit{LSCO -- LCB}; seed spread is largest under the more conservative 2D GNN evaluator.}
    \label{fig:seed_study}
\end{figure}

To assess how much results depend on the random seed for \emph{stochastic} codon optimization methods, we fix a single representative starting mAb (chosen near the median A280 of the dataset) and re-run each stochastic method across 10 seeds. Since the evaluators are deterministic given a sequence, the spread in Figure~\ref{fig:seed_study} reflects methods' stochasticity alone. Deterministic methods (original, natural, frequency, ICOR) are seed-inert and therefore omitted from the boxes. We report both LSCO variants (\textit{LSCO -- LCB}, \textit{LSCO -- LCB + MFE}) alongside the stochastic baselines CodonTransformer (temperature $0.4$) and GEMORNA, plus the LSCO random-substitution baseline.

Across seeds, the results in terms of performance generally hold with some caveats. Under the two 1D predictors, both LSCO variants stay above the stochastic baselines on predicted A280 for every seed, with bounded spread; \textit{LCB} variant is tightest, while \textit{LCB + MFE} shows slightly more variability and slightly lower absolute A280. Under the more conservative and less aligned 2D GNN, the boxes overlap more and seed spread is larger for all methods. Biophysical metrics are stable: both LSCO variants keep GC within the $50$--$60\%$ band on every seed, and \textit{LCB + MFE} consistently yields lower MFE than \textit{LCB}. These results are limited to a single starting mAb and characterize generator stochasticity rather than variation across antibodies.

\section{LSCO vs.\ Direct Input-Space Optimization (DISO)}
\label{app:diso}

DISO is an ablation of LSCO designed to isolate the contribution of the pretrained VMAE mRNA language model. The two methods share everything except \emph{where} the optimization happens: DISO reuses the exact same expression-predictor ensemble $\Psi$, MFE head $\Upsilon$, tokenizers, valid-codon mask, and candidate-selection gate as LSCO. The only thing that changes is the set of variables being optimized, and, as a direct consequence, how the naturalness prior is expressed. The empirical outcome of this comparison is reported in the main text (Table~\ref{tab:combined_sensitivity_ablation}): LSCO generally matches or exceeds DISO across evaluators, confirming that the pretrained latent space, and not merely the shared components, drives the gains.

\paragraph{What is being optimized (search geometry).}
LSCO optimizes a continuous latent code $z$ in the VMAE latent space, and turns it into a codon sequence by passing it through the pretrained decoder $D_\tau$ at every step. DISO removes the decoder from the loop entirely: it optimizes one set of \emph{per-position codon logits} directly in sequence space, one logit vector for each position of the heavy and light chains. In other words, LSCO moves through the smooth, learned manifold of the language model, while DISO moves through the raw codon-logit space with no generative model shaping the trajectory. Everything downstream — the temperature-controlled softmax, the valid-codon masking that keeps the sequence translating back to the target protein, the expression and MFE predictors — is identical.

\paragraph{Naturalness prior: latent anchoring vs.\ distribution matching.}
This is the component that changes most, and it changes because it \emph{has} to. In LSCO the naturalness prior lives in the latent space: we encode the BERT back-translated ``natural'' sequence to a reference code $z_{\text{nat}}$ and pull the current code toward it with a squared-distance penalty, $U_{\text{nat}}(z) = -\lVert z - z_{\text{nat}} \rVert^2$. DISO has no latent space, so this anchor cannot be defined the same way. Instead, DISO anchors the sequence directly in codon space: it builds a fixed per-position categorical distribution from the BERT-natural sequence (using a fixed reference temperature $\tau_{\text{BERT}}=0.5$) and, at each step, penalizes the Kullback--Leibler divergence between the current per-position codon distribution and this fixed reference, $\mathrm{KL}(p_{\text{current}} \,\Vert\, p_{\text{BERT}})$. For a two-chain antibody the term is the average of the heavy- and light-chain KL divergences. Conceptually both do the same job — keep candidates close to biologically plausible codon usage — but LSCO measures ``closeness'' as a distance between latent codes, whereas DISO measures it as a distance between probability distributions over codons. Setting $\lambda_{\text{nat}}=0$ removes the anchor in both methods.

\paragraph{Initialization.}
LSCO initializes the latent code $z$ from a standard Gaussian and lets the decoder map it to a starting sequence. DISO offers two starting points. The default (\texttt{init\_mode=bert}) warm-starts the logits at the BERT-natural codon for each position (a logit of $0$ for the natural codon and $-\delta$ for the others, with $\delta = 3.0$), so DISO begins from a natural sequence. The ``fairer-ablation'' variant (\texttt{init\_mode=random\_gauss}) instead draws logits from $\mathcal{N}(0, \sigma^2)$ with $\sigma = 0.5$ and no bias toward the natural codon, immediately suppressing invalid codons via the valid-codon mask so optimization still starts from a protein-consistent point. Importantly, the naturalness anchor target is \emph{always} the BERT-natural distribution regardless of where the logits are initialized — initialization only decides the starting point, not the anchor.

\paragraph{Optimization schedule and other hyperparameters.}
To keep the comparison as clean as possible, both methods share the same optimization budget and schedule: an identical decoding-temperature anneal from a high value (broad exploration) down to a low value (commit to a single codon per position), the same ramp on the constraint that enforces protein consistency, the same number of steps, learning rate, optimizer, and range of MFE weights. This is deliberate — because DISO differs from LSCO only in the search geometry (raw codon logits vs.\ the pretrained latent code) and the induced form of the naturalness anchor, matching all remaining settings ensures that any performance gap is attributable to those two factors rather than to differing training budgets. Table~\ref{tab:diso_hparams} lists the settings for both.

\begin{table}[t]
    \centering
    \resizebox{0.85\linewidth}{!}{%
    \begin{tabular}{lcc}
        \toprule
        & \textbf{LSCO} & \textbf{DISO} \\
        \midrule
        Optimized variable            & latent code $z$                 & per-position codon logits \\
        Generative decoder in loop    & Yes ($D_\tau$)                  & No \\
        Naturalness anchor            & latent $\ell^2$ distance $\lVert z - z_{\text{nat}}\rVert^2$ & KL to fixed BERT distribution \\
        Naturalness weight $\lambda_{\text{nat}}$ & $10^{-2}$ (0 to disable)               & $10^{-2}$ (0 to disable) \\
        Initialization                & $z \sim \mathcal{N}(0, I)$      & BERT-natural warm start (or $\mathcal{N}(0,0.5^2)$) \\
        Temperature schedule $\tau$   & $0.5 \rightarrow 0.03$ & $0.5 \rightarrow 0.03$ \\
        Optimization steps            & $3000$                          & $3000$ \\
        Learning rate                 & $5 \times 10^{-3}$                       & $5 \times 10^{-3}$  \\
        MFE weight $\lambda_{\text{MFE}}$ & $\{0, 0.1, 1\}$                       & $\{0, 0.1, 1\}$ \\
        Optimizer                     & AdamW                            & AdamW \\
        \bottomrule
    \end{tabular}%
    }
    \caption{Configuration differences between LSCO and DISO. All model components (expression ensemble, MFE head, tokenizers, valid-codon mask, candidate gate) are shared; only the search geometry and its downstream consequences differ.}
    \label{tab:diso_hparams}
\end{table}



\end{document}